\documentclass[letterpaper, 10 pt, journal, twoside]{IEEEtran}
\usepackage{amsmath,amsfonts}
\usepackage{array}
\usepackage[caption=false,font=normalsize,labelfont=sf,textfont=sf]{subfig}
\usepackage{textcomp}
\usepackage{stfloats}
\usepackage{url}
\usepackage{verbatim}
\usepackage{graphicx}
\usepackage{cite}

\usepackage{bm}
\usepackage{wrapfig}
\usepackage[ruled,vlined]{algorithm2e}
\usepackage[noend]{algpseudocode}
\usepackage{tabularx}
\usepackage{multirow, makecell}
\usepackage{amssymb}
\usepackage[table]{xcolor}
\usepackage{cancel}

\newcommand{\vx}{{\bf x}}

\newcommand{\vz}{{\bf z}}

\newcommand{\vu}{{\bf u}}

\newcommand{\vA}{{\bf A}}
\newcommand{\vR}{{\bf R}}

\newcommand{\vK}{{\bf K}}

\newcommand{\vF}{{\bf F}}

\newcommand{\vQ}{{\bf Q}}
\newcommand{\vB}{{\bf B}}

\newcommand{\vP}{{\bf P}}

\begin{document}

\title{TOAST: Trajectory Optimization and Simultaneous Tracking Using Shared Neural Network Dynamics}

\author{Taekyung Kim, Hojin Lee, Seongil Hong, and Wonsuk Lee
\thanks{Manuscript received: January 20, 2022; Revised: April 25, 2022; Accepted: June 1, 2022.}%
\thanks{This paper was recommended for publication by Editor C. Gosselin upon evaluation of the Associate Editor and Reviewers' comments. \textit{(Taekyung Kim and Hojin Lee are co-first authors.) (Corresponding author: Wonsuk Lee.)}}%
\thanks{The authors are with the Ground Technology Research Institute, Agency for Defense Development, Daejeon 34186, Republic of Korea {\tt\footnotesize \{ktk1501, hojini1117, hongsi, wsblues\}@add.re.kr}}%
\thanks{Digital Object Identifier (DOI): 10.1109/LRA.2022.3184769}
}
\markboth{IEEE Robotics and Automation Letters. Preprint Version. Accepted June, 2022}%
{Kim \MakeLowercase{\textit{et al.}}: TOAST: TRAJECTORY OPTIMIZATION AND SIMULTANEOUS TRACKING USING SHARED NEURAL NETWORK DYNAMICS}


\maketitle

\begin{abstract}
Neural networks have been increasingly employed in Model Predictive Controller (MPC) to control nonlinear dynamic systems. However, MPC still poses a problem that an achievable update rate is insufficient to cope with model uncertainty and external disturbances. In this paper, we present a novel control scheme that can design an optimal tracking controller using the neural network dynamics of the MPC, making it possible to be applied as a plug-and-play extension for any existing model-based feedforward controller. We also describe how our method handles a neural network containing history information, which does not follow a general form of dynamics. The proposed method is evaluated by its performance in classical control benchmarks with external disturbances. We also extend our control framework to be applied in an aggressive autonomous driving task with unknown friction. In all experiments, our method outperformed the compared methods by a large margin. Our controller also showed low control chattering levels, demonstrating that our feedback controller does not interfere with the optimal command of MPC. \footnote{Our video can be found at: \url{https://youtu.be/YQG0yHE5jWw}}
\end{abstract}


\section{INTRODUCTION}
\IEEEPARstart{O}{nline} trajectory optimization, also known as Model Predictive Control (MPC), has provided promising results in numerous robotic applications. The most popular current methods for controlling high-dimensional nonlinear systems involve gradient-based MPC, which was established in the context of differential dynamic programming \cite{abbeel_application_2007, tassa_synthesis_2012, tassa_control-limited_2014}. However, this approach poses the inherent limitation that the algorithm relies on a quadratic approximation for cost functions and requires smooth dynamics. Furthermore, it is extremely difficult to include state constraints in the optimization process of gradient-based MPC.

As a remedy, sampling-based approaches that do not require linear and quadratic approximations of dynamics and cost functions have been developed. The key advantage of sampling-based MPC is that it is capable of solving non-convex optimization problems, allowing engineers to easily encode high-level robot behaviors using cost function clipping \cite{ kuwata_real-time_2009, williams_aggressive_2016}. With the benefit that the algorithm can directly use the dynamics model without linearization, neural networks that have been proven effective for approximating the nonlinear transition models are increasingly employed in sampling-based MPC \cite{nagabandi_dexterous_2019, williams_information-theoretic_2018}.

\IEEEpubidadjcol

Both of these MPC algorithms (called feedforward controllers) optimize an open-loop control sequence and execute this sequence until the next optimization update. However, an achievable update rate of the MPC is insufficient to cope properly with model uncertainty and external disturbances. This is because the optimization process must incorporate a number of trajectory rollouts along the future time horizon to reach the required performance level. In practice, a simply designed local feedback compensator is often used to correct for trajectory errors on the fastest time scales \cite{murray_optimization-based_2009}. The fundamental problem is that such compensators decouple the feedback and feedforward control domains, resulting in decreased efficiency due to the lack of consideration for the feedforward optimization process. As an alternative, the iterative Linear Quadratic Regulator (iLQR) \cite{todorov_generalized_2005}, which is also known as a Sequential Linear Quadratic (SLQ) solver \cite{sideris_efficient_2005}, was proposed to be used in an online setting for deriving the optimal feedforward trajectory and the feedback gains simultaneously by solving a single optimal control problem \cite{tassa_synthesis_2012, neunert_fast_2016}. However, such schemes have only been considered in gradient-based settings and they are not scalable to sampling-based algorithms. In contrast to gradient-based MPC, the optimal feedback controller for sampling-based MPC has not been fully addressed by prior work.

One particular strategy to deal with this issue is to apply iLQR on top of the sampling-based MPC for designing an ancillary feedback policy \cite{williams_robust_2018, williams_information-theoretic_2018}. However, the feedback gain term in iLQR is produced to push the control sequence to the locally optimal trajectory during the line search optimization procedure. The feedback gain becomes negligible when the sequence converges after recursive backward and forward passes. Nonetheless, it is possible to obtain a considerable feedback gain by intentionally reducing the iterations, but it is misleading to refer to this as an ideal optimal control gain. Finally, all the feedback strategies mentioned above have not yet considered a proper extension to neural network dynamics.

\begin{figure*}[t]
\centering
\includegraphics[width=0.95\textwidth]{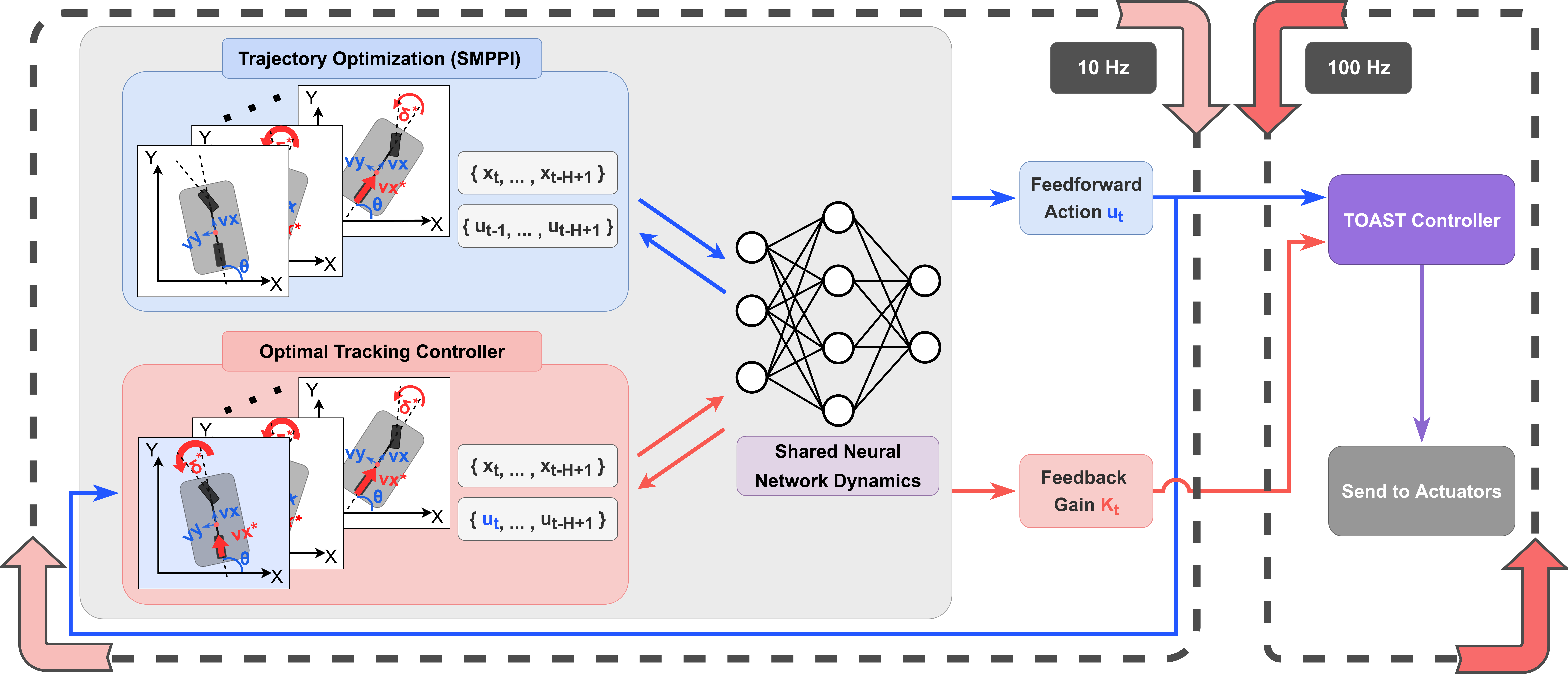}
\caption{Overview diagram of our TOAST control framework when applied to an aggressive driving task of an Unmanned Ground Vehicle (UGV). The components related to the feedforward controller and the feedback controller are depicted in blue and red, respectively.}
\label{fig:flow_chart}
\end{figure*}

In this paper, we propose a novel approach that combines trajectory optimization and optimal tracking control via sharing neural network dynamics, which is applicable to both gradient-based and sampling-based MPC. Since our method is allowed to use the neural network transition models of the system, it can be applied as a plug-and-play extension to previously completed model-based control tasks. We also describe how our method handles a neural network containing history information, in which such a strategy is often employed for capturing time-varying or higher-order effects. We evaluated our algorithm using classical control benchmarks in which crosswinds are modeled as external disturbances. In addition, we demonstrated the feasibility of our idea by performing a challenging autonomous driving task under varying road friction conditions. The proposed method called Trajectory Optimization and Simultaneous Tracking (TOAST) is illustrated in Fig.~\ref{fig:flow_chart}.
\section{METHODOLOGY}
In this paper, we consider the problem of designing a generic control framework that includes MPC and a corresponding optimal tracking controller that both employ the same learned dynamics. While the idea is general and applicable to any MPC algorithm, such as iLQG \cite{tassa_synthesis_2012} and Differential Dynamic Programming (DDP) \cite{tassa_control-limited_2014}, we use Smooth Model Predictive Path Integral (SMPPI) control \cite{kim_smooth_2022} as the feedforward controller throughout this paper.

\subsection{Smooth Model Predictive Path Integral Controller}
This section provides a brief overview of the SMPPI approach. Consider a standard discrete time dynamic system in which we denote the state at time $t$ as $\vx_t \in \mathbb{R}^{n}$ and the action as $\vu_t \in \mathbb{R}^{m}$. A Model Predictive Path Integral (MPPI) controller, which is one of the most successful sampling-based approaches in recent years, draws parallel control sequence samples $U = \{\vu_0, \vu_1, \dots \vu_{T-1}\}$ with a fixed time horizon $T$ \cite{williams_aggressive_2016}. The optimal control trajectory $U^*$ is then obtained using the information theoretic interpretation and importance sampling. The parallel samples are evaluated with the state cost function $c(\vx)$ using a Graphics Processing Unit (GPU). However, chattering innately occurs in the optimized control sequence due to the nature of sampling-based algorithms. SMPPI is proposed to attenuate the chattering without extrinsic smoothing filters. The main idea is to decouple the control space and action space. Noisy sampling is performed on the control sequences and then the integral operation is applied on them to obtain action sequences while smoothing chattering.
\subsection{Designing a Simultaneous Optimal Tracking Controller}
Online trajectory optimization methods, as well as SMPPI, have shown effective performance in various dynamic systems. However, there remains a problem: the computational demand is too high for a real-world system to process in the required time frame. In an MPC-controlled system, the same feedforward command should be maintained for a certain amount of processing time. Control errors induced by external disturbances or model uncertainty might be difficult to respond to in such a system.

We propose an optimal feedback controller that minimizes the control errors that occur while the MPC calculates the next feedforward command. This ancillary feedback controller does not interfere with the optimal command of MPC, but rather takes the feedforward command as a reference and performs tracking control that reduces the error at a faster rate. Our method does not need to form a new error function to track the feedforward action, nor does it need to learn separate neural network dynamics. It calculates the optimal feedback gain adaptively by utilizing the same learned dynamics from MPC. Here, we emphasize that the feedback controller discussed in this paper is distinct from the low-level plant controller, which manages actuators based on the control input and is often referred to as a black-box.

Consider a discrete time nonlinear system dynamics model:
\begin{equation} 
    \vx_{t+1} = \vF(\vx_t, \vu_t).
    \label{Equation:dynamics}
\end{equation}

An optimal action trajectory found by MPC is given as $U^* = \{\vu^*_0, \vu^*_1, \dots \vu^*_{T-1}\}$ and corresponding state trajectory is given as $X^* = \{\vx^*_0, \vx^*_1, \dots \vx^*_{T-1}\}$. Linearize (\ref{Equation:dynamics}) by using first-order Taylor series expansion:
\begin{equation} 
\begin{aligned} 
    \vx_{t+1} 
    & \approx \vF(\vx_{t}^{*}, \vu_{t}^{*}) \, + \\
    & \underbrace{\left.\frac{\partial \vF}{\partial \vx_{t}}\right|_{\vx_{t}=\vx_{t}^{*}}}_{\vA_t} \left(\vx_{t}-\vx_{t}^{*}\right)
    + \underbrace{\left.\frac{\partial \vF}{\partial \vu_{t}}\right|_{\vu_{t}=\vu_{t}^{*}}}_{\vB_t} \left(\vu_{t}-\vu_{t}^{*}\right).
\end{aligned}
\label{Equation:taylor}
\end{equation}
As the definition of (\ref{Equation:dynamics}), (\ref{Equation:taylor}) is equivalent to:
\begin{equation} 
    \vx_{t+1} - \vx_{t+1}^{*} \approx \vA_t \left(\vx_{t}-\vx_{t}^{*}\right)
    + \vB_t \left(\vu_{t}-\vu_{t}^{*}\right).
\end{equation}
Here, we define new variables $\vz_t = \vx_t - \vx_t^*$ and $\bm{\nu}_t = \vu_t - \vu_t^*$. Then the system dynamics can be expressed as:
\begin{equation} 
    \vz_{t+1} = \vA_t \, {\vz_t} + \vB_t \, {\bm{\nu}_t}.
    \label{Equation:new_dynamics}
\end{equation}
Assume a quadratic cost function:
\begin{equation} 
    J = \sum_{t=0}^{\infty} \vz_{t}^{\top} \, {\vQ} \, \vz_{t} + \bm{\nu}_{t}^{\top} \, {\vR} \, \bm{\nu}_{t},
    \label{Equation:objective}
\end{equation}
where $\vQ \geq 0, \vR > 0$. Then, it seems to be formulated as a standard LQR problem \cite{bansal_learning_2016}. Here, we use the infinite-time objective function for computational efficiency. The resulting errors from this simplification would be sufficiently compensated by the online optimization strategy of MPC. It is worth mentioning that the dynamics near the optimal trajectory can be interpreted as being locally linear because the discretized time frame of MPC would be short enough. As a result, a linear feedback controller such as LQR can be adopted for this problem.

One should note that the tracking controller will run faster than the MPC. The discretized time period in (\ref{Equation:new_dynamics}) is denoted by $\Delta{t}_1$, and the one in (\ref{Equation:objective}) is denoted by $\Delta{t}_2$. Converting the system dynamics (\ref{Equation:new_dynamics}) to have faster sampling period $\Delta{t}_2$ results in:
\begin{equation} 
\begin{aligned}
    \tilde{\vz}_{t+1}
        & =\left(1-\frac{\Delta t_{2}}{\Delta t_{1}}\right) \tilde{\vz}_{t}+\frac{\Delta t_{2}}{\Delta t_{1}}\left(\vA_t \tilde{\vz}_{t}+\vB_t \tilde{\bm{\nu}}_{t}\right) \\
        & = \underbrace{\left[\frac{\Delta t_{2}}{\Delta t_{1}} \vA_t+\left(1-\frac{\Delta t_{2}}{\Delta t_{1}}\right) I_{n}\right]}_{\tilde{\vA}_t} \tilde{\vz}_{t}+ \underbrace{\frac{\Delta t_{2}}{\Delta t_{1}} \vB_t}_{\tilde{\vB}_t} \, \tilde{\bm{\nu}}_{t} \\
        & = {\tilde{\vA}_t} \, \tilde{\vz}_{t}+ {\tilde{\vB}_t} \, \tilde{\bm{\nu}}_{t},
\end{aligned}
\label{Equation:converted_dynamics}
\end{equation}
where we use notation $\tilde{\left( \cdot \right)}$ to denote variables that are discretized into a shorter sampling period. Notice that the formulation of the converted dynamics in (\ref{Equation:converted_dynamics}) is that of a common linear interpolation.

In the discrete time linear quadratic control problem of (\ref{Equation:objective}) and (\ref{Equation:converted_dynamics}), we can solve for $\tilde{\bm{\nu}}_t$ to yield the optimal linear feedback policy using the discrete Riccati equation \cite{anderson_optimal_2007}:
\begin{equation} 
    \tilde{\bm{\nu}}_t = -{\vK_t} \, \tilde{{\vz}}_t,
\label{Equation:feedback_policy}
\end{equation}
where $\vK_t$ is feedback gain:
\begin{equation} 
\begin{aligned}
    \vK_t & = ({\vR} + \tilde{\vB}_t^{\top}{\vP_t}\tilde{\vB}_t)^{-1}\tilde{\vB}_t^{\top}{\vP_t}\tilde{\vA}_t, \\
    \vP_t & = \tilde{\vA}_t^{\top} \vP_t \tilde{\vA}_t - \\
    & \quad \tilde{\vA}_t^{\top} \vP_t \tilde{\vB}_t(\vR+\tilde{\vB}_t^{\top} \vP_t \tilde{\vB}_t)^{-1} \tilde{\vB}_t^{\top} \vP_t \tilde{\vA}_t+\vQ .
\end{aligned}
\label{Equation:solve_gain}
\end{equation}

\subsection{Trajectory Optimization and Simultaneous Tracking}
\begin{algorithm}[tb]
\SetKwInOut{Input}{Given}
\Input{
        $\vF_{\bm{\theta}}$: Neural network dynamics model \\
        $\vQ, \vR$: Feedback control parameters \\}   
\While{task not completed}{
$\vx_0^{*} \leftarrow \text{StateEstimation}();$ \\
$\vu_0^{*} \leftarrow 
\text{SMPPI} \left( \vF_{\bm{\theta}}, \, \vx_0^{*} \right);$ \\
$\vx_1^{*} \leftarrow \vF_{\bm{\theta}}\left( \vx_0^{*}, \,  \vu_0^{*} \right);$ \\
$\vA_t, \vB_t \leftarrow \text{AutoGrad}\left( \vx_1^{*}, \, \left[ \vx_0^{*}, \, \vu_0^{*} \right] \right);$ \\
$\tilde{\vA}_t, \, \tilde{\vB}_t \leftarrow \text{ConvertDynamics}\left( \vA_t, \, \vB_t \right);$\Comment{Eq. (\ref{Equation:converted_dynamics})} \\
$ \vK_t \leftarrow \text{SolveLQR} \left( \tilde{\vA}_t, \, \tilde{\vB}_t, \,  \vQ, \, \vR \right);$\Comment{Eq. (\ref{Equation:solve_gain})} \\
\While{processing time of SMPPI}{
    $\tilde{\vx}_t \leftarrow \text{StateEstimation}();$ \\
    $\tilde{\vx}_t^{*} \leftarrow \text{Interpolation}\left( \vx_0^{*}, \, \vx_1^{*} \right);$ \Comment{Eq. (\ref{Equation:interpolation})} \\
    $\tilde{\vu}_t \leftarrow \vu_0^{*} - \vK_t \left( \tilde{\vx}_t - \tilde{\vx}_t^{*} \right);$ \\
    $\text{SendToController}(\tilde{\vu}_t);$ \\
}
}
\caption{TOAST \label{Algorithm:TOAST}} 
\end{algorithm} 

In the previous section, we demonstrated that the system dynamics for trajectory optimization can also be utilized by a trajectory following feedback controller. We now describe how the two controllers in our proposed control framework, called Trajectory Optimization and Simultaneous Tracking (TOAST), operate harmoniously. In addition, we will address practical issues that may arise during implementation.

Expanding (\ref{Equation:feedback_policy}) we obtain:
\begin{equation} 
    \tilde{\vu}_t - \tilde{\vu}_t^{*}  = -{\vK_t} \, \left(\tilde{\vx}_t - \tilde{\vx}_t^{*}\right).
\end{equation}
Note that $\tilde{\vu}_t^{*}$ is the optimal feedforward command from MPC. Here, $\tilde{\vu}_t^{*}$ can be replaced with $\vu_0^{*}$, as it will maintain the same command during the processing time $\Delta{t}_1$ of MPC. Then, we have:
\begin{equation} 
    \tilde{\vu}_t  =  \vu_0^{*} -{\vK_t} \, \left(\tilde{\vx}_t - \tilde{\vx}_t^{*}\right).
\label{Equation:final_action}
\end{equation}
Since $\tilde{\vx}_t$ can be obtained by measuring directly from the sensor or only through state estimation algorithms, such as an extended Kalman filter, it can be updated at a much faster refresh rate than the feedforward commands.

Let $t_0$ be the instant at which the feedforward command $\vu_0^{*}$ is received from MPC. And let $t$ be the instant at which the feedback controller receives the state $\tilde{\vx}_t$. Then, $\tilde{\vx}_t^{*}$ in (\ref{Equation:final_action}) is synchronously updated via linear interpolation:
\begin{equation} 
    \tilde{\vx}_t^{*} = \left(1 - \frac{t - t_0}{\Delta t_{1}}\right) \vx_0^{*} + \left(\frac{t - t_0}{\Delta t_{1}}\right) \vx_1^{*} .
\label{Equation:interpolation}
\end{equation}
From the aforementioned derivations, we can derive a generic control scheme that includes an optimal feedforward controller and a locally linear optimal feedback policy that adapts to the former. If there are no external disturbances and sensor noises, $\left(\tilde{\vx}_t - \tilde{\vx}_t^{*}\right)$ will remain nearly zero, and the feedforward command $\vu_0^{*}$ will take precedence during $\Delta t_{1}$. Conversely, if the state of the system deviates from the expected trajectory [as given in (\ref{Equation:interpolation})], the feedback command will start to operate to complement the feedforward command and keep the system on the desired trajectory.

Since the feedback controller utilizes the same dynamics $\vF$ as the feedforward controller, our control scheme can be applied very effectively to existing control tasks where the system dynamics models are highly complex and non-linear. In recent studies of MPC \cite{abbeel_application_2007, williams_aggressive_2016} and model-based reinforcement learning \cite{nagabandi_dexterous_2019}, for instance, the dynamics models are approximated by neural network. Our feedback controller can use the learned dynamics ($\vF_{\bm{\theta}}$) directly to solve optimal tracking problems without additional resources.

Note that the reference state at the next time step $\vx_1^{*}$ in (\ref{Equation:interpolation}) can not be acquired instantly during feedforward trajectory optimization, since sampling-based MPC indirectly optimizes the action trajectory $U^*$ rather than the state trajectory $X^*$. Meanwhile, we implement the linearization to obtain the Jacobians of neural network dynamics in the Pytorch \texttt{autograd} automatic differentiation package \cite{paszke_automatic_2017}. For automatically computing the gradients of $\vF_{\bm{\theta}}$ with respect to $\vx_t$ and $\vu_t$, one step forward propagation with the model (\ref{Equation:dynamics}) must be done first in order to obtain $\vx_{t+1}$. We can take advantage of this result to interpolate the current reference state to circumvent additional computation. The overall algorithm of our TOAST framework is shown in Algorithm~\ref{Algorithm:TOAST}.

\subsection{Extension to an Aggressive Autonomous Driving Task \label{subsection:extension}}
\begin{figure}[t] 
\centering
\subfloat[]{\includegraphics[width=0.23\textwidth]{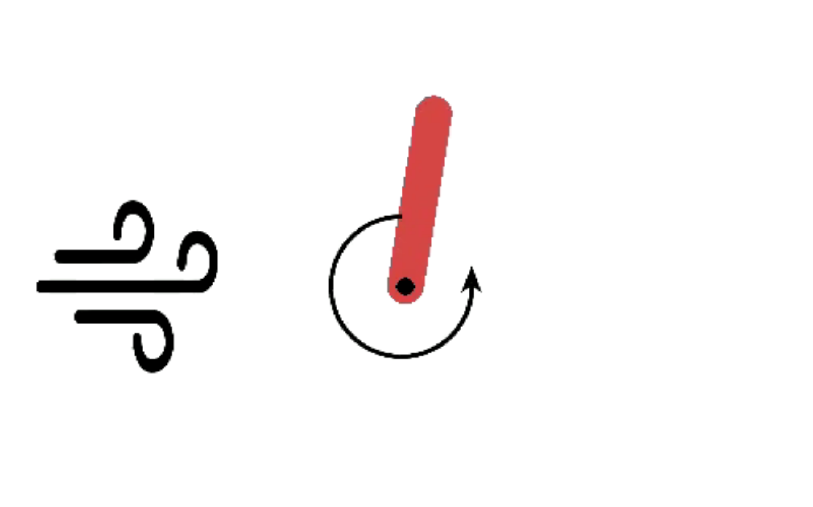}} \hspace*{0.01\textwidth}
\subfloat[]{\includegraphics[width=0.23\textwidth]{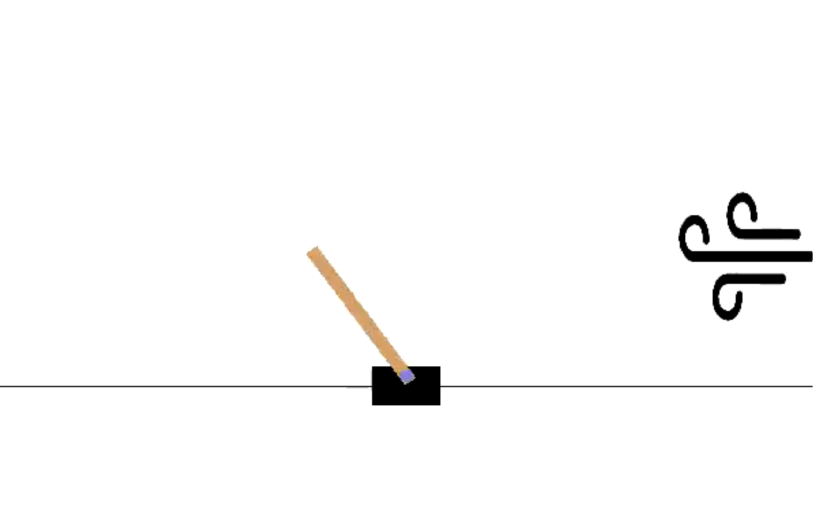}} \hfill
\subfloat[]{\includegraphics[width=0.48\textwidth]{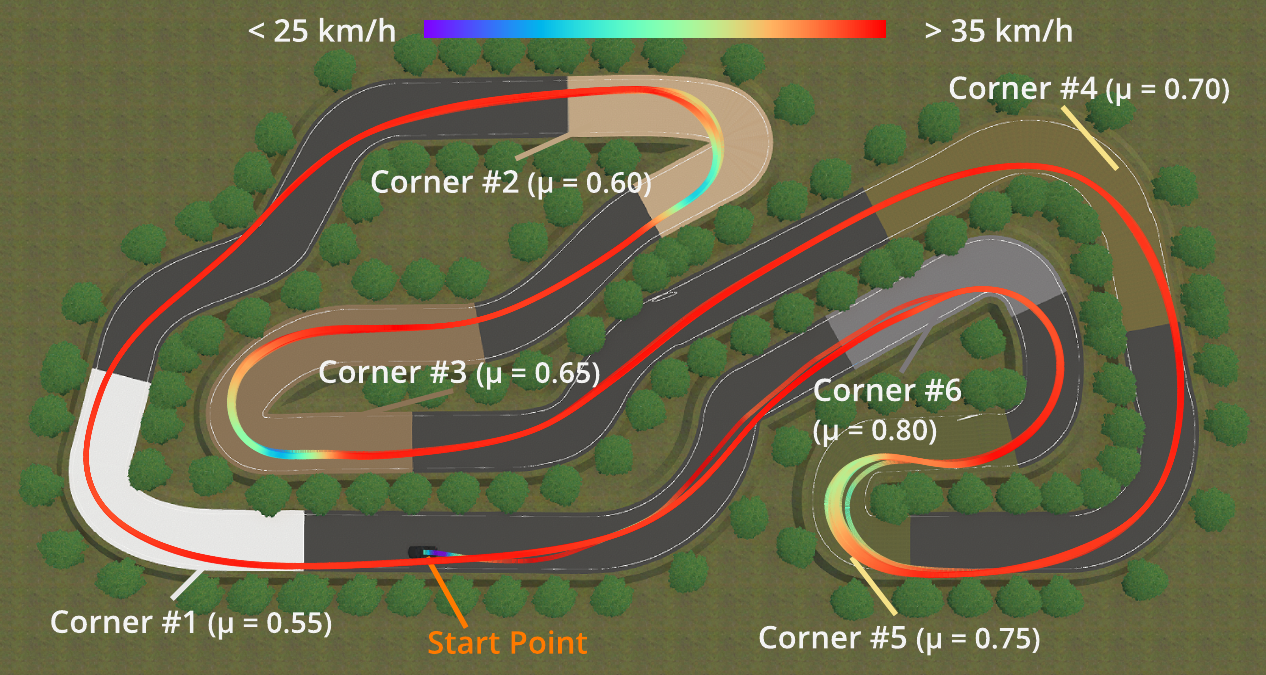}}  
\caption{We evaluate our algorithm using classical control benchmarks and aggressive autonomous driving. Crosswinds are modeled in (a) Pendulum, and (b) Cartpole as external disturbances. (c) A high-fidelity vehicle simulator is used for the autonomous driving experiments. The vehicle trajectory is also visualized using our control framework for driving 10 laps around the track in a counter-clockwise direction.}
\label{fig:exp_env}
\end{figure}

Our research goal is to control an UGV to drive aggressively under challenging conditions, such as sharp corners and unknown surfaces with high and low friction. The vehicle will slide on low-friction corners, causing tire lateral forces to enter the highly nonlinear zone. Despite longitudinal and lateral accelerations being robust features for estimating future states of sliding vehicles, such measurements from real-world sensors are highly noisy. One effective solution for this problem is to provide a history of the vehicle's velocity states and control inputs to the neural network \cite{spielberg_neural_2019, spielberg_neural_2021}. Following our earlier study \cite{kim_smooth_2022} where the results were also promising, we determine the history length as $4$. Because the history length is relatively short, we employ a fully-connected neural network for computational efficiency. Note that other types of model architectures, such as Recurrent Neural Networks (RNN), can also be utilized if the Jacobians of the model can be obtained. However, ablation studies on model architecture selection are beyond the scope of this paper.

Then, the approximated model can be expressed in the following form:
\begin{equation}
    \vx_{t+1} = \vF_{\bm{\theta}}\left( \vx_t, \,  \vu_t, \, \vx_{t-1}, \,  \vu_{t-1}, \, \dots \vx_{t-H+1}, \,  \vu_{t-H+1} \right) ,
\label{Equation:history_dynamics}
\end{equation}
where $H$ is denoted as the length of the history. Since (\ref{Equation:history_dynamics}) does not follow a general form of dynamics, direct linearization is not applicable. Substituting this dynamics model into the Taylor linearization (\ref{Equation:taylor}) yields:
\begin{equation} 
\begin{aligned} 
    & \vx_{t+1} - \vx_{t+1}^{*} 
    \approx \vA_t \left(\vx_{t}^{*} + \delta{\vx_t}-\vx_{t}^{*}\right)
    + \vB_t \left(\vu_{t}^{*}+ \delta{\vu_t}-\vu_{t}^{*}\right) \\
    & \qquad \qquad \qquad \qquad \qquad \quad \dots \\
    & + \left.\frac{\partial \vF_{\bm{\theta}}}{\partial \vx_{t-H+1}}\right|_{\vx_{t-H+1}^{*}} \cancelto{0}{\left(\vx_{t-H+1}^{*}+\delta{\vx_{t-H+1}}-\vx_{t-H+1}^{*}\right)} \\
    & + \left.\frac{\partial \vF_{\bm{\theta}}}{\partial \vu_{t-H+1}}\right|_{\vu_{t-H+1}^{*}} \cancelto{0}{\left(\vu_{t-H+1}^{*}+\delta{\vu_{t-H+1}}-\vu_{t-H+1}^{*}\right)} ,
\end{aligned}
\end{equation}
where we can replace $\left( \vx \right)^* + \delta\left( \vx \right)$ with $\left( \vx \right)$, and replace $\left( \vu \right)^* + \delta\left( \vu \right)$ as $\left( \vu \right)$. The state and control deviations $\delta\left( \vx \right)$ and $\delta\left( \vu \right)$ before the current time step $t$ were presented in the past and they should be zero at time $t$. As a result of causality, linearization of the dynamics with historical encoding can be addressed as a current-state-dependent system.

In a normal vehicle state estimation setting, the state $\vx$ in approximated dynamics $\vF_{\bm{\theta}}$ consists of the longitudinal velocity $v_x$, the lateral velocity $v_y$, and the yaw rate $r$. However, the original purpose of the tracking controller was not to follow the command and state at velocity level, but to follow them at position level. We, therefore, augment the state space with dynamic state variables $\vx_d = \left(v_x, v_y, r\right)^{\top}$ and kinematic state variables $\vx_k$ such that:
\begin{equation}
    \mathbf{x}=\left(\begin{array}{c}
    \mathbf{x}_{d} \\
    \mathbf{x}_{k}
    \end{array}\right),
\end{equation}
where $\vx_k$ consists of x-position $p_x$, y-position $p_y$, and yaw angle $\theta$ on the global coordinate frame. The next kinematic state can be updated by the current dynamic state:
\begin{equation}
\mathbf{x}_{k}(t+1)=\mathbf{x}_{k}(t)+\left(\begin{array}{c}
\cos (\theta) v_{x}-\sin (\theta) v_{y} \\
\sin (\theta) v_{x}+\cos (\theta) v_{y} \\
r
\end{array}\right) \Delta t_{1} .
\label{Equation:explicit}
\end{equation}
The augmented full state is then updated via the neural network $\vF_{\bm{\theta}}$ and the explicit kinematic function (\ref{Equation:explicit}). Linearization can still be done using the automatic differentiation method. The overall control architecture, as it was applied to an autonomous driving task, is depicted in Fig.~\ref{fig:flow_chart}.

\section{EXPERIMENTS}

\begin{figure*}[t]
\centering
\includegraphics[width=0.98\textwidth]{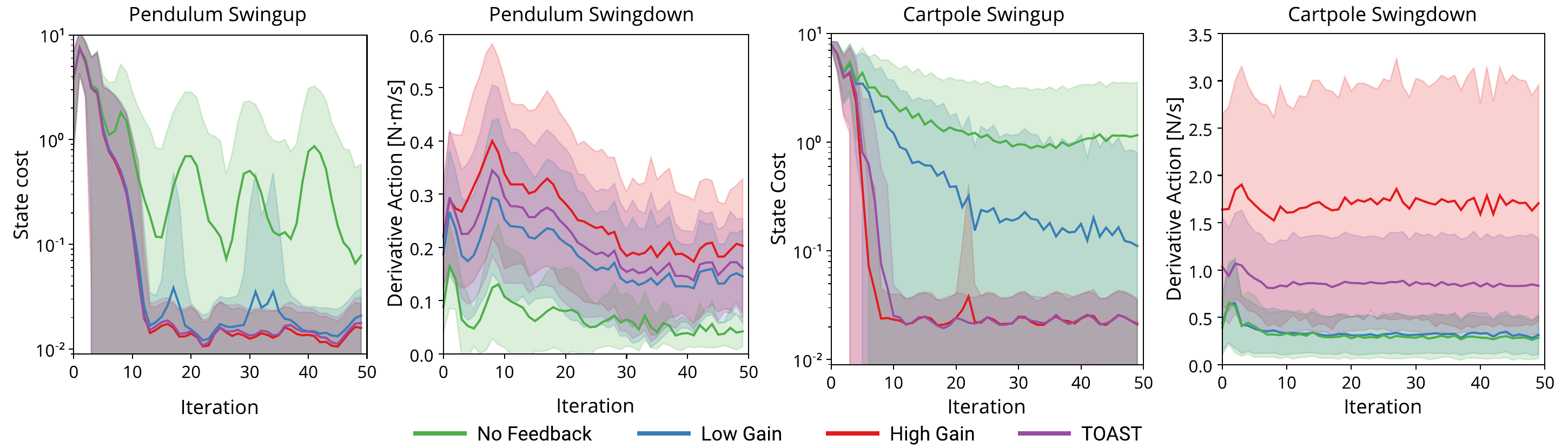}
\caption{Experimental results in four control tasks. The derivative actions are quantified using the L2 norm.}
\label{fig:gym_exp}
\end{figure*}

For performance comparison, we implement four feedback controllers based on the same feedforward controller, SMPPI. These are: 1) no feedback controller (abbreviated as ``No Feedback"), 2) a feedback controller with a low hand-tuned gain matrix (abbreviated as ``Low Gain"), 3) a feedback controller with a high hand-tuned gain matrix (abbreviated as ``High Gain"), and 4) our TOAST controller. Note that the TOAST's minimum and maximum gains observed throughout the experiments are set to ``Low Gain" and ``High Gain", respectively. We perform a grid search to find the best control parameters showing the most successful performance for each task. The discretized time step for feedforward $\Delta{t}_1$ is set to $0.1$ s, while the feedback $\Delta{t}_2$ is set to $0.01$ s. In the experiments, we use a neural network to approximate the system dynamics.

\subsection{Pendulum}
We first evaluate the performances of the compared controllers using classical control benchmarks: Pendulum and Cartpole. To emulate external disturbances, one can consider adding perturbations in the transition dynamics \cite{mankowitz_robust_2020} or adding a noise in action \cite{tessler_action_2019}. We use a more intuitive method by modeling a continuously varying crosswind. The crosswind applied to the pole's surface is converted to imposed torque. This environment is visualized in Fig.~\ref{fig:exp_env}a.

We designed the neural network dynamics with two fully-connected hidden layers, each of which has $32$ neurons. The model predicts the residual difference between the current state and the next state: \begin{equation}
\begin{bmatrix}
    \theta_{t+1}\\
    \dot{\theta}_{t+1}
\end{bmatrix} = 
\begin{bmatrix}
    \theta_t\\
    \dot{\theta}_t
\end{bmatrix} +
\vF_{\bm{\theta}}\left( \begin{bmatrix}
                    \sin(\theta_t)\\
                    \cos(\theta_t)\\
                    \dot{\theta}_t\\
                    \vu_t
                    \end{bmatrix}
\right).
\end{equation}
We designed two distinct tasks: 1) swing-up the pole just as in other control studies, and 2) swing-down the pole to maintain a static state. The swing-down task is designed to assess the chattering level of the controller when the system is in a stable condition. The running cost function of the swing-up task is formulated as:
\begin{equation} 
    c(\vx \equiv [\theta, \dot{\theta}]^{\top}) = \theta^2 + 0.1 \, {\dot{\theta}^2} .
\end{equation}
Likewise, the cost function of the swing-down task is formulated to keep the pole downward against external disturbances:
\begin{equation} 
    c(\vx \equiv [\theta, \dot{\theta}]^{\top}) = (\theta+ \pi)^2 + 0.1 \, {\dot{\theta}^2} .
\end{equation}

\begin{figure}[t]
\centering
\includegraphics[width=0.95\linewidth]{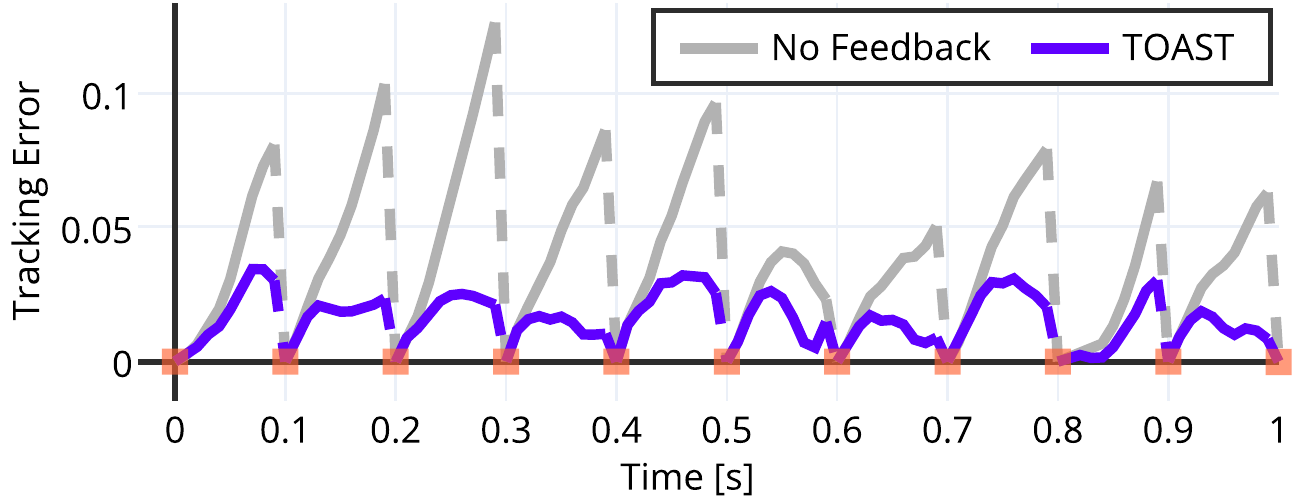}
\caption{Visualization of tracking errors during a Pendulum swing-up task. The feedforward controller optimizes a new trajectory every $0.1$ s. At that time, the tracking error is updated to zero (red squares). TOAST is effectively regulating the trajectory deviations caused by external disturbances.}
\label{fig:tracking_err}
\end{figure}

While only MPC was activated, a dataset was collected online with $1000$ random bootstrap data and the model was trained with the dataset every $50$ time-steps until the pole was balanced successfully in an upright position. After training, the feedback controller was activated for the experiments. For each task, the dynamics were trained with $10$ different fixed random seeds, and the controllers were tested at $50$ random locations for each of the pre-trained dynamics. In the first half of the $500$ tests, random winds were applied, whereas sinusoidal winds were applied in the second half. Each controller was tested under exactly the same conditions except for the feedback controller. We analyzed the state costs for the swing-up task, and the derivative actions for the swing-down task to quantify the degree of chattering (shown in Fig.~\ref{fig:gym_exp}). The tracking errors of ``No Feedback" and TOAST during a Pendulum swing-up test are depicted in Fig.~\ref{fig:tracking_err}.

For SMPPI, $1000$ trajectories with $15$ time-steps were sampled for trajectory optimization. For TOAST, the control parameters were set as follows: $\vQ = \text{Diag}(1, \, 0.1)$, $\vR = \text{Diag}(0.001)$.
\subsection{Cartpole}
The majority of Cartpole's experimental settings follow the Pendulum. We modify the common Cartpole environment to provide a continuous action space with a range of $(-10, 10)$ N, in which the force is applied to the cart horizontally (shown in Fig.~\ref{fig:exp_env}b). We use neural network dynamics with the same architecture as Pendulum:
\begin{equation}
\begin{bmatrix}
    \dot{x}_{t+1}\\
    \dot{\theta}_{t+1}
\end{bmatrix} = 
\begin{bmatrix}
    \dot{x}_{t}\\
    \dot{\theta}_{t}
\end{bmatrix} +
\vF_{\bm{\theta}}\left(
    \begin{bmatrix}
        \sin(\theta_t)\\
        \cos(\theta_t)\\
        \dot{\theta}_t\\
        \vu_t
    \end{bmatrix}
\right),
\end{equation}
where the velocity of the cart $\dot{x}$ can be derived using only the model inputs. Then, the kinematic variables can be obtained using Euler integration as described in Section~\ref{subsection:extension}:
\begin{equation}
\begin{bmatrix}
    x_{t+1}\\
    \theta_{t+1}
\end{bmatrix} = 
\begin{bmatrix}
    x_t\\
    \theta_t
\end{bmatrix} +
\begin{bmatrix}
    \dot{x}_t\\
    \dot{\theta}_t
\end{bmatrix}
\Delta t .
\end{equation}

The running cost function of the swing-up task is formulated as:
\begin{equation} 
    c(\vx \equiv [x, \dot{x}, \theta, \dot{\theta}]^{\top}) = 50 \, x^2 + 10 \, \dot{x}^2 + 10 \, \theta^2 + {\dot{\theta}^2} .
\end{equation}
The cost function of the swing-down task follows the same method as for Pendulum.

For TOAST, the control parameters were set as follows: $\vQ = \text{Diag}(1, \, 1, \, 10, \, 10)$, $\vR = \text{Diag}(0.001)$. The signs of the hand-tuned gains for ``Low Gain" and ``High Gain" were switched depending on whether the pole was placed in the upper half-plane or the lower half-plane. This is because the action in Cartpole, unlike in Pendulum, is applied to the cart horizontally.

The experimental results of four tasks are shown in Fig.~\ref{fig:gym_exp}. The results show that both ``High Gain" and TOAST could keep the pole upward against the continuously varying external disturbances. MPC without a feedback controller failed to control the systems in most cases. ``Low Gain" performed better than ``No Feedback", but still failed to achieve the desired performance level. On the other hand, ``Low Gain" and ``No Feedback" showed low action chattering levels in the swing-down tasks, as expected. TOAST also demonstrated a much lower chattering level than ``High Gain", suggesting that our proposed feedback controller does not interfere with MPC when the tracking errors are assumed to be trivial.

\subsection{Aggressive Autonomous Driving}
We also evaluated the performances of different controllers in a high-fidelity vehicle simulator, the IPG CarMaker. It has been widely used to validate precisely nonlinear vehicle dynamics \cite{acosta_tire_2018, dixit_trajectory_2020, bae_curriculum_2021}. We built a race track with a length of $1016$ m, two moderate curves, and four sharp curves (see Fig.~\ref{fig:exp_env}c). A Volvo XC90 is used as the control vehicle. Unlike the electric vehicles used in recent learning-based aggressive autonomous driving studies \cite{williams_aggressive_2016, kabzan_learning-based_2019}, our vehicle used an automatic transmission for gear-shifting. Therefore, we employ desired speed ($v_{des}$), instead of throttle, as a high-level controller's input and let a low-level plant controller manage the throttle and brake. Thus, the control input $\vu$ consists of steering angle $\delta$ and desired speed $v_{des}$.

\subsubsection{Training Neural Network Vehicle Dynamics}

Our fully-connected neural network approximates the nonlinear vehicle dynamics. Note that only the dynamic state variables are considered as input to the neural network. We denote $\left|\vx_{d}\right|$ and $\left|\vu\right|$ as the size of the controller's state and action. The history of states and actions, which has a size of $(\left|\vx_{d}\right| + \left|\vu\right|) \cdot H $, is forward propagated to the neural network. Then, the output of the network predicts the residual difference between the current state and the next state, which has the size of $\left|\vx_{d}\right|$. The network was implemented as an MLP with four hidden layers. We applied a hyperbolic tangent (tanh) activation function, and the network was trained with the mean squared error loss function and Adam optimizer.

We collected a human-controlled driving dataset, with a data rate of $10$ Hz, to train our network by expanding the methods used in our previous work \cite{bae_curriculum_2021}. We found that the dataset should comprise three types of distinct maneuvers in order for the neural network to accurately represent the vehicle dynamics for various friction conditions: 
\begin{enumerate}
    \item Zig-zag driving at low speeds ($20-25$ km/h) on the race track, in both clockwise and counter-clockwise directions. Each driving in a different direction was treated as a separate maneuver.
    
    \item High speed driving on the race track in both directions, trying to maintain $40$ km/h as much as possible.
    
    \item Sliding maneuvers at the friction limits on flat ground, in combinations of acceleration and deceleration with various steering angles.
\end{enumerate}
The above five maneuvers were done with seven different friction coefficients: $[0.4, 0.5, 0.6, 0.7, 0.8, 0.9, 1.0]$. These $35$ maneuvers were logged for two minutes each, obtaining a total of $70$ minutes of driving data. We divided the data into $70 \, \%$ for training and $30 \, \%$ for testing after randomizing the data to break temporal correlations. The trained model was saved when it showed the best result for test error. Then, it was evaluated with the validation dataset. We collected the validation data on the race track in both directions. The default friction coefficient was set to $0.8$ and the friction coefficients of the six curves were set to $[0.95, 0.85, 0.75, 0.65, 0.55, 0.45]$, respectively, which were previously not included in the training data.

The test and validation errors of our neural network are shown in Table~\ref{table:errors}. Root Mean Square Error (RMSE) is denoted as $\mathbf{E}_{RMS}$ and the max error is denoted as $\mathbf{E}_{max}$. The results show that our neural network can make accurate one-step predictions under a variety of driving conditions.

\begin{table}[ht]
\renewcommand\arraystretch{1.2}
\captionsetup{justification=centering}
\caption{Test and validation errors of the neural network vehicle dynamics.}
\label{table:errors}
\begin{center}
\begin{tabularx}{1.0\columnwidth}{c|cc|cc|cc|}
& \multicolumn{2}{c|}{$v_x$ [m/s]} & \multicolumn{2}{c|}{$v_y$ [m/s]} & \multicolumn{2}{c|}{$r$ [rad/s]} \\ \cline{2-7}
& $\mathbf{E}_{RMS}$ & $\mathbf{E}_{max}$ & $\mathbf{E}_{RMS}$ & $\mathbf{E}_{max}$ & $\mathbf{E}_{RMS}$ & $\mathbf{E}_{max}$  \\ \cline{1-7}
\textbf{Test} & 0.0273 & 0.3940 & 0.0178 & 0.3180 & 0.0147 & 0.3455 \\
\textbf{Val.} & 0.0220 & 0.3753 & 0.0114 & 0.1127 & 0.0095 & 0.1413 \\
\end{tabularx}
\end{center}
\vspace*{-0.15in}
\end{table}

\subsubsection{Experimental Setup}
We designed the state-dependent cost function $c(\vx)$ in MPC to have the following form:
\begin{equation} \label{eq:cost}
\begin{aligned}
    c(\vx) & = \alpha_1{\text{Track}(\vx)} + \alpha_2{\text{Speed}(\vx)} + \alpha_3{\text{Slip}(\vx)}, \\
    \text{Track}(\vx) & = (0.9)^t\,{10000 \, \textbf{M}(p_x,p_y)} , \\
    \text{Speed}(\vx) & = (v_x - v_{ref})^2 , \\
    \text{Slip}(\vx) & = \sigma^2 + 10000I \, (\{\left | \sigma \right | > 0.2\}) ,
\end{aligned}
\end{equation}
where $I$ is an indicator function. $\textbf{M}(p_x,p_y)$ in track cost is the two-dimensional cost map value at the global position $(p_x, p_y)$. Thanks to the sampling-based derivation of SMPPI, we can provide an impulse-like penalty in the cost function. The speed cost is a simple quadratic cost to achieve the reference vehicle speed $v_{ref}$. The slip cost imposes both soft and hard costs to discourage slip angle in the trajectory $( \sigma = -\text{arctan}  ( \frac{v_y}{\lVert v_x \rVert} ) )$. The trajectory expected to have a slip angle greater than $0.2$ rad (approximately $11.46 \, ^{\circ}$) will be penalized since it has the potential to make the vehicle unstable.

For SMPPI, the number of parallel samples was $10000$ and the number of time steps was $35$. For TOAST, the control parameters were set as follows:
\begin{equation}
\begin{aligned}
\vQ &= \text{Diag}
\begin{pmatrix}
5, & 5, & 0.01, & 0.05, & 0.05, & 0.025 
\end{pmatrix}, \\
\vR &= \text{Diag}
\begin{pmatrix}
0.4, & 0.1
\end{pmatrix} .
\end{aligned}
\end{equation}

\begin{figure}[t]
\centering
\includegraphics[width=0.99\linewidth]{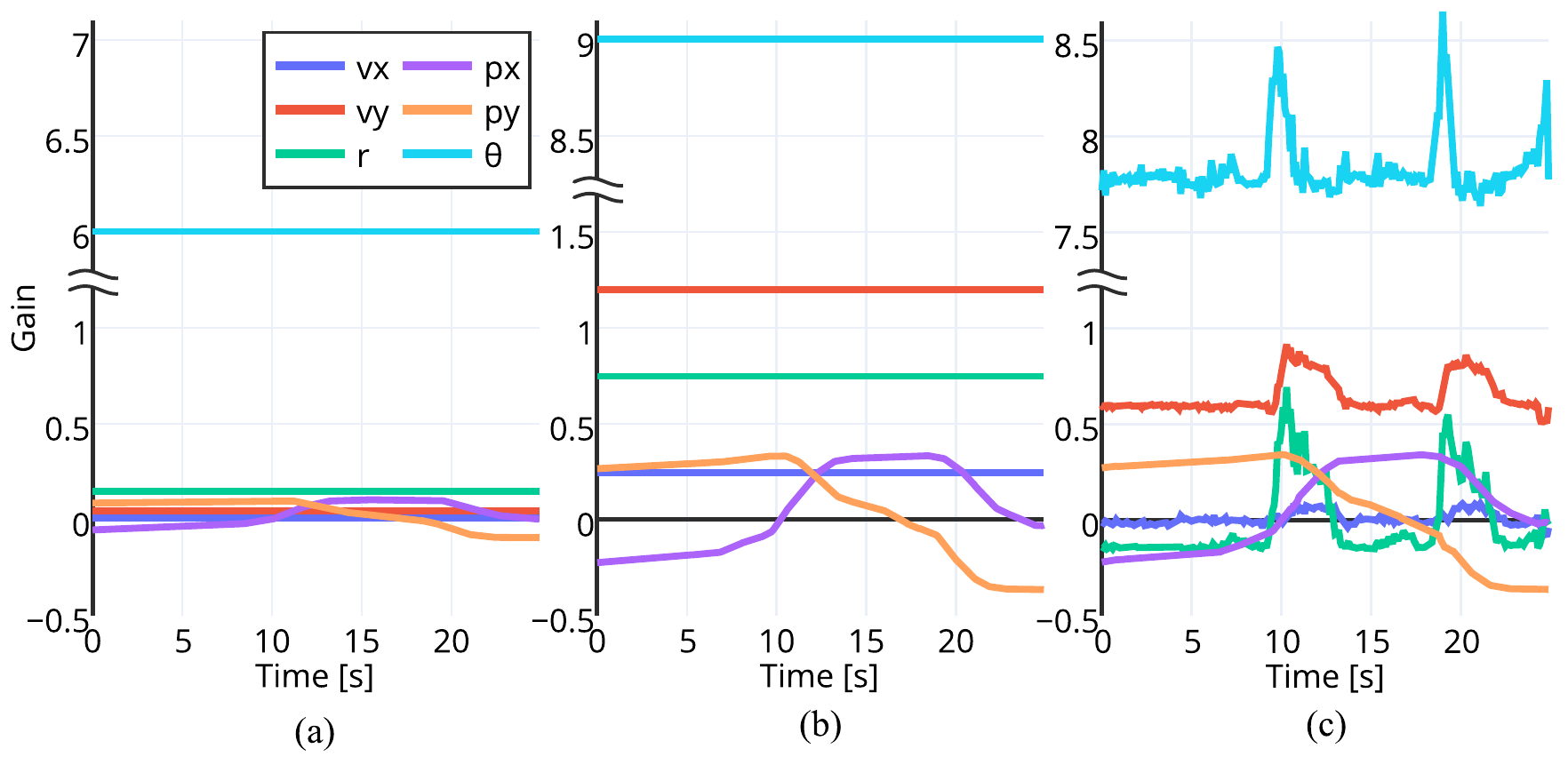}
\caption{Visualization of the time-varying feedback gains with different controllers while they are driving the same section. (a) ``Low Gain", (b) ``High Gain", and (c) TOAST. For clarity, only the gains corresponding to the steering angle are visualized. We applied rotation on the hand-tuned gains of $p_x$ and $p_y$ from global to vehicle coordinate frame. TOAST adaptively changes the gains according to the time-varying dynamic characteristics of the system.}
\label{fig:gain_plot}
\end{figure}
\begin{figure}[t]
\centering
\includegraphics[width=0.95\linewidth]{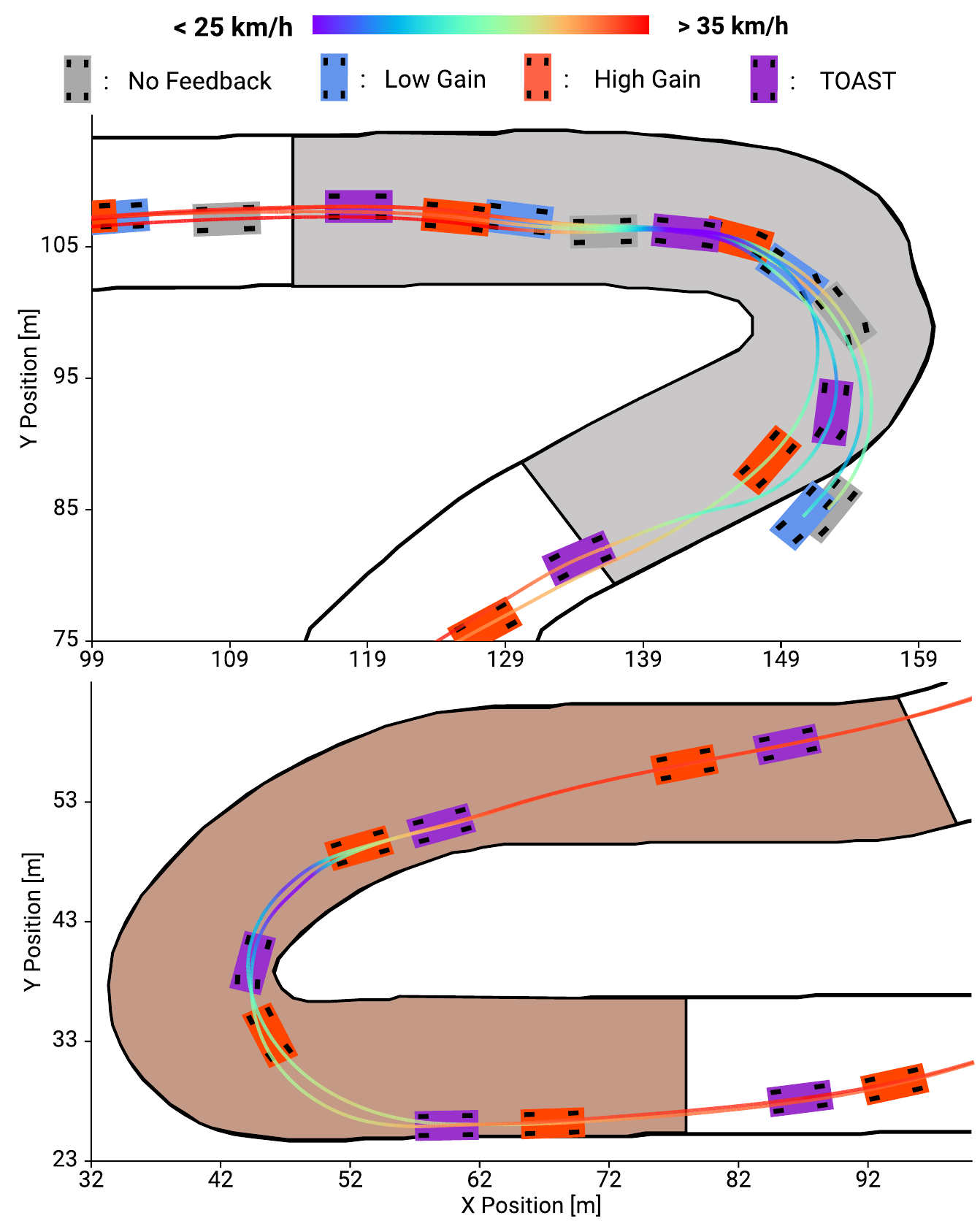}
\caption{Vehicle trajectories of the compared controllers. The friction coefficient of \textit{Corner \#2} (in gray) is 0.60 and \textit{Corner \#3} (in brown) is 0.65.}
\label{fig:trajectory_compare}
\end{figure}

\subsubsection{Experimental Results\label{subsubsection:result}}
We evaluated the control performance of the four controllers on the race track. The goal of the controllers was to complete whole laps in a clockwise direction.

The race track was adjusted to be more challenging to drive than it was when the training dataset was collected. The friction coefficients $\mu$ were modified to have values of $[0.55, 0.60, 0.65, 0.70, 0.75, 0.80]$ at the corners, respectively, and to have a default of $1.0$. During ten laps around the course, we measured average lap times on the six corners at a reference speed of $40$ km/h. When the vehicle left the track, it was placed at the starting point and started a new lap. All of the methods used the same pre-trained model and no further training was given throughout the experiments. The results are shown in Table~\ref{table:time_took}. 

\begin{table}[ht]
\renewcommand\arraystretch{1.2}
\captionsetup{justification=centering}
\caption{Average lap times on the six corners of different control methods. The minimum speed (in [km/h]) and maximum slip angle (in [$^{\circ}$]) at each corner are also analyzed with our method. The success rates (SR) are also displayed.}
\label{table:time_took}
\begin{center}
\begin{tabular}{c|cccccc|c}
Lap time [s] & \#1 & \#2 & \#3 & \#4 & \#5 & \#6 & SR \\\hline
No Feedback      & 7.8 & 9.4 & 12.2 & 8.7 & 7.0 & 11.9  & 1/10   \\
Low Gain & 7.9 & 8.9 & 12.2 & 8.9 & 6.9 & 12.2  & 3/10  \\
High Gain      & 8.3 & 8.7 & 12.2 & 10.0 & 7.6 & 13.6  & 10/10  \\
\rowcolor{gray!50} TOAST     & 7.7 & 7.8 & 11.8 & 9.5 & 8.3 & 9.4  & 10/10  \\  \hline \hline
\rowcolor{gray!50} Min. Speed      & 33.2 & 24.4 & 22.6 & 33.5 & 26.7 & 26.7  &  \\ 
\rowcolor{gray!50} Max. Slip      & 3.4 & 7.2 & 6.8 & 2.8 & 7.4 & 4.2  &  \\ 
\end{tabular}
\end{center}
\vspace*{-0.1in}
\end{table}

``No Feedback" and ``Low Gain" barely got through \textit{Corner \#2} and \textit{Corner \#3}, which are the most challenging sections of the track. They completed the full course with only one and three out of ten laps, respectively. The learned dynamics predicted that the vehicle would slide greatly if it did not slow down due to the low friction surfaces. Therefore, SMPPI generated trajectories that applied brakes first, then steered the vehicle and slowly accelerated it to pass the corners. However, the vehicle failed to follow the planned trajectory and understeering occurred due to model uncertainty. In contrast, ``High Gain" and ``TOAST" completed entire laps in $100\%$ of the trials. However, the results show that TOAST has a faster lap time than ``High Gain" in most cases. This is because if the feedback controller keeps a fixed high gain all the time, it frequently interferes with the control sequence of the MPC and thus loses the optimality of the solution. Our controller computes the optimal feedback gain for each optimal control sequence based on the contextual information encoded in the neural network dynamics. As a result, a low gain can be maintained in situations where feedback compensation is not greatly required. We show the time-varying feedback gains of the compared controllers in Fig.~\ref{fig:gain_plot}. For TOAST, the overall average speed during ten laps was $33.4$ km/h and the maximum slip angle was $7.4 \, ^{\circ}$. We visualize the vehicle trajectories on \textit{Corner \#2} and \textit{Corner \#3} in Fig.~\ref{fig:trajectory_compare}.

\begin{wrapfigure}{R}{0.5\linewidth}
\centering
\includegraphics[width=0.99\linewidth]{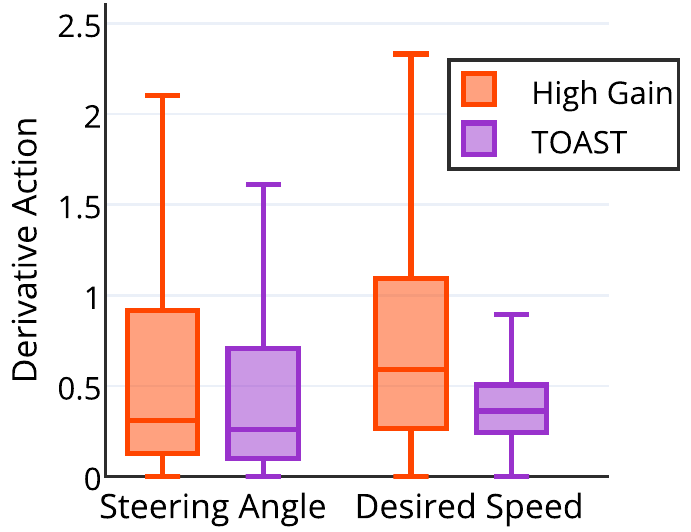}
\caption{Derivative action for each control variable. They are quantified by the L2 norm.}
\label{fig:du}
\end{wrapfigure}

We also analyzed the degree of chattering on control values with ``High Gain" and TOAST. It is a well-known fact that rapid changes in the action commands are a burden to the actuators. In the straight sections of the track with high friction coefficients, the vehicle is stable and the requirement for a feedback controller becomes negligible. Therefore, we analyzed the derivative actions during $10$ laps excluding the corners. The results are shown in Fig.~\ref{fig:du}. For ``High Gain", the average derivative actions of the steering angle ($\delta$) and the desired speed ($v_{des}$) were $1.31$, and $0.81$, respectively. On the other hand, those of TOAST were $0.96$ and $0.46$. The results demonstrate that the adaptive manner of our proposed method can alleviate chattering.
\section{CONCLUSION}
We presented a novel control scheme that combines online trajectory optimization and optimal tracking control using the same learned dynamics. We described how to convert a single dynamics model so that it could be used for controllers on two different time scales. We also explained how to handle dynamics models containing history information for our method. The performance of the proposed method was evaluated using two control benchmarks, demonstrating that our ancillary feedback controller is capable of regulating tracking errors without interfering with the optimal feedforward control sequence. We also extended our work to an autonomous aggressive driving task, suggesting its scalability. In all experiments, our method outperformed the compared methods by a large margin. Our controller also showed control chattering levels that were $24 - 43 \%$ lower than for the controller with high hand-tuned gains in the driving task.

\bibliographystyle{IEEEtran}
\typeout{}
\bibliography{references.bib}

\end{document}